\documentclass[
]{ceurart}

\usepackage{listings}
\usepackage{todonotes}
\usepackage{graphicx}
\usepackage{booktabs}
\usepackage{multicol, multirow}
\usepackage{amsmath}
\usepackage{amssymb}
\usepackage{wrapfig}
\usepackage{xcolor,colortbl}
\usepackage{url}
\usepackage{array} 

\newcommand{\no}{\cellcolor{lightgray}}

\begin{document}

\copyrightyear{2025}
\copyrightclause{Copyright for this paper by its authors.
  Use permitted under Creative Commons License Attribution 4.0
  International (CC BY 4.0).}

\conference{CLEF 2025 Working Notes, September 9 -- 12 September 2025, Madrid, Spain}

\title{NLPnorth @ TalentCLEF 2025:\\Comparing Discriminative, Contrastive, and Prompt-Based Methods for Job Title and Skill Matching} 


\author[1]{Mike Zhang}[%
orcid=0000-0003-1218-5201,
email=jjz@cs.aau.dk,
url=https://jjzha.github.io/,
]
\cormark[1]
\fnmark[1]
\address[1]{Aalborg University, Copenhagen, Denmark}

\author[2]{Rob van der Goot}[%
orcid=0009-0003-1999-4156,
email=robv@itu.dk,
url=https://robvanderg.github.io/,
]
\fnmark[1]
\address[2]{IT University of Copenhagen, Copenhagen, Denmark}

\cortext[1]{Corresponding author.}
\fntext[1]{These authors contributed equally.}

\begin{abstract}
Matching job titles is a highly relevant task in the computational job market domain, as it improves e.g., automatic candidate matching, career path prediction, and job market analysis. 
Furthermore, aligning job titles to job skills can be considered an extension to this task, with similar relevance for the same downstream tasks. 
In this report, we outline NLPnorth's submission to TalentCLEF 2025, which includes both of these tasks: Multilingual Job Title Matching, and Job Title-Based Skill Prediction. 
For both tasks we compare (fine-tuned) classification-based, (fine-tuned) contrastive-based, and prompting methods. 
We observe that for Task A, our prompting approach performs best with an average of 0.492 mean average precision (MAP) on test data, averaged over English, Spanish, and German. 
For Task B, we obtain an MAP of 0.290 on test data with our fine-tuned classification-based approach. 
Additionally, we made use of extra data by pulling all the language-specific titles and corresponding \emph{descriptions} from ESCO for each job and skill. 
Overall, we find that the largest multilingual language models perform best for both tasks.
Per the provisional results and only counting the unique teams, the ranking on Task A is 5$^{\text{th}}$/20 and for Task B 3$^{\text{rd}}$/14.
\end{abstract}

\begin{keywords}
  computational job market analysis \sep
  NLP for Human Resources \sep
  job title matching \sep
  job-skill matching \sep
  classification \sep
  contrastive learning \sep
  prompting \sep
  large language models
\end{keywords}

\maketitle

\section{Introduction}
The dynamic and rapidly evolving nature of labor markets, primarily driven by technological advancements~\cite{lasi2014industry,schwab2017fourth,eu2021industry,eloundou2023gpts}, global migration patterns~\cite{autor2013growth}, digitization~\cite{autor2003skill}, and economic shifts, has significantly increased the availability of detailed job advertisement data across various recruitment and employment platforms. These platforms actively leverage job postings to attract qualified candidates, generating rich and structured datasets that are highly valuable for labor market analysis~\cite{brynjolfsson2011race,brynjolfsson2014second,balog2012expertise}. Consequently, there has been substantial growth in research related to Natural Language Processing (NLP) applications for Human Resources~\cite{balog2012expertise,otani2025naturallanguageprocessinghuman} and Computational Job Market Analysis~\cite{senger-etal-2024-deep,zhang2024computational}.

Specific research efforts in this area include skill extraction from job postings, a task crucial for identifying current workforce needs, emerging job roles, and skill shortages. Existing methods for skill extraction range from traditional rule-based and pattern-matching approaches to advanced machine learning and deep learning techniques, using both supervised and unsupervised methods~\cite{sayfullina2018learning,bhola-etal-2020-retrieving,khaouja2021unsupervised,zhang-etal-2022-skillspan,zhang-jensen-plank:2022:LREC,zhang2022skill,green-maynard-lin:2022:LREC,gnehm-bhlmann-clematide:2022:LREC,nguyen2024rethinking,herandi2024skill,vasquez2024hardware,zhang2024nnose,kavas2025multilingual}.

Complementary research focuses on skill classification and matching, which aims to align extracted skills to other similar skills or to existing taxonomies such as ESCO~\cite{le2014esco}. This research area frequently explores innovative methodologies, such as leveraging self-supervised learning techniques, transformer-based language models, and semantic embeddings for skill representation and matching~\cite{giabelli2020graphlmi,degroot2021job,yazdanian2021radar,decorte2022design,zhang-jensen-plank:2022:LREC,clavie2023large,decorte2024skillmatchevaluatingselfsupervisedlearning,magron-etal-2024-jobskape,gavrilescu2025techniques}. Similarly, job title classification and matching to, e.g., SOC~\cite{elias2010soc2010} or ISCO~\cite{elias1997occupational} have been explored extensively, addressing challenges related to standardizing diverse job titles, which vary significantly across organizations and regions. Techniques include supervised classification, semantic matching algorithms, and transformer-based models such as JobBERT~\cite{decorte2021jobbert}, which facilitate matching and categorization of job titles~\cite{javed2015carotene,javed2016towards,javed2017large,green-maynard-lin:2022:LREC,retyk2024melo,liu2024job}. These skills and job titles are also being further classified into their respective taxonomical counterparts to be used for further labor market demand analysis~\cite{malherbe2016bridge,sibarani2017ontology}.

Furthermore, research in career path prediction leverages historical job market data to anticipate future career trajectories, facilitating career counseling and workforce planning~\cite{decorte2023career,senger-etal-2025-karrierewege}. Lastly, significant attention has been devoted to mapping jobs and skills onto standardized occupational and skill taxonomies, aiding systematic labor market research and policy-making decisions~\cite{esco-2022,clavie2023large,vrolijk2023enhancingplmperformancelabour,zhang2024eljob,rosenberger2025careerbert}.

In this report, we investigate \emph{multilingual job–title matching} (TalentCLEF Task A) and \emph{job–skill matching} (Task B). Both tasks require mapping free-form labour-market text onto a structured knowledge base, yet they differ in language coverage, label granularity, and training-data volume. We show example annotations for both tasks in Table~\ref{tab:examples}. We systematically compare three paradigms that dominate current NLP practice: (1) Classification, which treats ranking as binary relevance prediction and fine-tunes a multilingual encoder with cross-entropy loss. (2) Contrastive learning, which learns task-specific sentence embeddings via InfoNCE and relies on cosine ranking at inference. (3) Prompting, which directly exploits instruction-tuned LLM embedders in a zero-shot setting, requiring no task-specific updates.

\begin{table}
\centering
\caption{Example annotations for the job title ``cider master'' examples from the training data. For task B there are 330 skill matches, we only show a selection here.}
\label{tab:examples}
\begin{tabular}{l l}
\toprule
Task A & Task B \\
\midrule
    master cider maker & analyse apple juice for cider production \\ 
cider production manager & analyse samples of food and beverages\\
 cider manufacture manager & communicate with external laboratories\\
 cider maker & display spirits\\
 & ... \\
 \bottomrule
\end{tabular}
\end{table}

We build all models on top of ESCO job titles and skills provided by the shared task organizers, for the contrastive models, we additionally exploit corresponding ESCO descriptions. Our experimental results show that contrastive learning excels on multilingual title matching, whereas discriminative fine-tuning leads on job–skill prediction.  Surprisingly, zero-shot prompting with a 7B parameter model performs close to supervised systems on Task A, highlighting the rapid progress of instruction-tuned LLMs for Computational Job Market Analysis.

\paragraph{Contributions:}
\begin{itemize}
    \itemsep0em
    \item We present the \emph{first side-by-side comparison} of discriminative, contrastive, and prompting paradigms for both job–title and job–skill matching.
    \item We introduce a \emph{unified ESCO-derived training corpus} (titles, alternative labels, and multilingual descriptions) and release preprocessing scripts to facilitate future research.\footnote{Code is released at: \url{https://github.com/jjzha/talentclef-nlpnorth}}
    \item We achieve \textbf{5\textsuperscript{th}} place out of 20 teams on Task A and \textbf{3\textsuperscript{rd}} out of 14 on Task B by carefully tuning model size, negative-sampling strategy, and inference prompts, demonstrating that \emph{model choice should be task-specific} rather than one-size-fits-all.
\end{itemize}

\section{The Tasks}
\subsection{TalentCLEF Task A} 
This task challenges participants to build multilingual systems that, given a job title, rank related titles from a knowledge base~\cite{gasco2025overview}. Provided resources include:
\begin{itemize}
    \itemsep0em
    \item \textbf{Training Data}: 15,000 labeled pairs of related job titles per language (English, Spanish, German) support cross-lingual training.
    \item \textbf{Development Data}: Participants receive 100 manually annotated samples per language, each with a query job title and related titles. A language-specific knowledge base is also provided for ranking. This data also includes Chinese.
    \item \textbf{Test Data}: Participants receive 5,000 job titles, with evaluation based on a gold-standard subset of 100 titles per language. Titles include hidden annotations for industry and gender, allowing bias assessment alongside ranking performance. This data also includes Chinese.
\end{itemize}

\subsection{TalentCLEF Task B}
This task challenges participants to build models that, given a job title, return related skills from a knowledge base. 
All data is in English.
\begin{itemize}
    \itemsep0em
    \item \textbf{Training Data}: 2,000 job titles with associated professional skills, sourced from real job descriptions and semi-automatically curated for accuracy.
    \item \textbf{Development Data}: 200 job titles with related skills normalized to ESCO terminology. A skills knowledge base is also provided.
    \item \textbf{Test Data}: 500 job titles for which participants must predict related skills using the provided knowledge base.
\end{itemize}

\section{Methodology}
We compare three main categories of approaches to tackle the tasks. Below, we will outline each approach, and which variations within the approach we evaluated. For each approach, we first describe the setup for job title matching, and in the final paragraph describe how the model is adapted to perform skill matching. We train all models on the concatenation of the data for all languages.

\subsection{Classification}
\paragraph{Training Data.} We reformulate the ranking task as binary classification. In Task A every provided positive instance consists of a query title and one related title. We create negative instances by pairing the same query with randomly sampled, unrelated titles. We experiment with positive–to–negative ratios of 1:1, 1:2, and 1:5 and select the best ratio (1:2) on the English development data. At prediction time, we use the output of the softmax to create a ranking (as opposed to using the binary labels directly). For classification, we only use TalentCLEF's training data and no extra descriptions as in the contrastive learning approach (Subsection \ref{sec:contrastive}).

\paragraph{Training.} We train with MaChAmp~\cite{van-der-goot-etal-2021-massive}, which places a single linear layer on top of a multilingual encoder and optimizes cross-entropy loss. We fine-tune several multilingual models given their domain representativeness and high performance on other classification tasks: \texttt{escoxlm-r}~\cite{zhang2023escoxlmr}, \texttt{m-e5-large-instruct}~\cite{zhang-etal-2024-e5}, \texttt{m-e5-large}~\cite{zhang-etal-2024-e5}, \texttt{mdeberta-v3-base}~\cite{he2021debertav3}, and \texttt{par-m-mpnet-base-v2}~\cite{reimers-2020-multilingual-sentence-bert}.

\paragraph{Hyperparameters.}
We started from the default hyperparameters from MaChAmp 0.4.2, which are a learning rate of 0.0001, batch size of 32, 20 epochs, and a slanted-triangular learning rate scheduler~\cite{howard-ruder-2018-universal}. In our initial runs, we found early convergence, so we experimented with a lower learning rate and a smaller amount of epochs.  We empirically saw similar performance with only 3 epochs, but the lower learning rate was not beneficial. Hence, we used all default settings except the number of epochs.

\paragraph{Adapting to Task B.} For Task B we keep the architecture unchanged and simply replace related titles with related skills. Negatives are sampled from the entire skill vocabulary, the optimal negative ratio for taskB was 1:1.


\subsection{Contrastive Learning}\label{sec:contrastive}

\paragraph{Training data.} We construct pairs from ESCO. For each occupation we create (i) preferred title $\rightarrow$ description and (ii) preferred title $\rightarrow$ alternative title. Negatives come from the remaining descriptions or titles.

\paragraph{Training.} 
We further fine-tune sentence embedding models with InfoNCE~\cite{oord2018representation}.
Given a batch of $N$ aligned pairs $(u_i,v_i)$, we treat every other $v_j$ ($j\neq i$) as a hard negative for $u_i$ and vice versa:
\begin{equation}
\ell_i = -\log \frac{\exp(\mathrm{sim}(u_i, v_i))}{\sum_{j=1}^{N} \exp(\mathrm{sim}(u_i, v_j))}
\label{eq:infonce}
\end{equation}
We use cosine similarity for $\mathrm{sim}(\cdot)$. The overall loss is the average of the $\ell_i$.

\paragraph{Hyperparameters.} We systematically vary the number of in-batch negatives $k\!\in\!\{1,2,5,10,15,16,20,32\}$, batch size $\!\in\!\{16,32,64\}$, and learning rate $\!\in\!\{1\times10^{-4},5\times10^{-5},2\times10^{-5},2\times10^{-6}\}$. The optimal configuration uses $k\!=\!16$, batch size 32, and learning rate $2\times10^{-6}$.

\paragraph{Adapting to Task~B.} We build three pair types: (i) job $\rightarrow$ skill, (ii) job $\rightarrow$ alt\_skill, and (iii) alt\_job $\rightarrow$ skill. We sample negatives from the complementary pool (all skills or all jobs, respectively).

\subsection{Prompting}

Finally, we test instruction-tuned LLM embedders in a zero-shot setting. Specifically, we embed text with \texttt{multilingual-e5-large-instruct} (560M), \texttt{Linq-Embed-Mistral} (7.11B) and \texttt{gte-Qwen2-7B-instruct} (7.61B). In this setup, we embed the query with a task description prefix, and we embed the candidate jobs/skills separately. Then, we rank candidates by cosine similarity. We use the following prefixes:
\begin{quote}\small
Task A: \texttt{``Given a job title, find the most relevant job titles.''}
\end{quote}
\begin{quote}\small
Task B: \texttt{``Given a job title, find the most relevant skills.''}
\end{quote}
\vspace{-.5cm}

\begin{table}
    \centering
    \caption{\textbf{Results Task A.} MAP scores for classification, similarity, and prompting-based methods. For the test predictions, we select the best performing model from each category.}
    \label{tab:a}
    \small
    \setlength{\tabcolsep}{0.5em} 
    \begin{tabular}{ll llllllllll}
    \toprule
    & & \multicolumn{5}{c}{\textbf{Validation}} & \multicolumn{5}{c}{\textbf{Test}} \\
    \midrule
    \multicolumn{11}{c}{\textbf{Classification} (fine-tuned)} \\
                                                  & \#P        & en        & de        & es        & zh        & \textbf{avg}           & en & de & es & zh & \textbf{avg} \\
    \midrule
                        TalentCLEF Baseline         & ?M         & 0.499 	& 0.377    & 0.284 	  & 0.437 	 & 0.399  & 0.408  & 0.348 	& 0.324 & 0.380 & 0.365 \\
                        \midrule
                        escoxlm-r                   & 561M       & 0.550    & 0.376    & 0.421    & 0.505    & 0.463 &\no&\no&\no&\no&\no\\
                        mdeberta-v3-base            & 276M       & 0.550    & 0.414    & 0.429    & 0.505    & 0.475\footnotemark[2] &\no&\no&\no&\no&\no\\
                        m-e5-large-instruct         & 560M       & 0.555    & 0.401    & 0.421    & 0.512    & \textbf{0.472}  & 0.490 & 0.421 & 0.406 & 0.444 & 0.440 \\
                        m-e5-large                  & 560M       & 0.548    & 0.369    & 0.416    & 0.500    & 0.458 &\no&\no&\no&\no&\no\\
                        par-m-mpnet-base-v2         & 278M       & 0.530    & 0.312    & 0.418    & 0.496    & 0.439 &\no&\no&\no&\no&\no\\
    \midrule
    \multicolumn{11}{c}{\textbf{Contrastive} (fine-tuned)} \\
    \midrule   
                                            & \#P       & en        & de        & es        & zh            & \textbf{avg}      & en & de & es & zh & \textbf{avg} \\
    \midrule
    escoxlm-r                               & 561M      &  0.578    & 0.405     & 0.473     & 0.516         & 0.493  &\no&\no&\no&\no&\no\\
    m-e5-large                              & 560M      &  0.612    & 0.446     & 0.481     & 0.539         & \textbf{0.520}   & 0.511 & 0.446 & 0.484 & 0.477 & 0.480 \\
    par-m-mpnet-base-v2                     & 278M      &  0.549    & 0.368     & 0.448     & 0.480         & 0.461  &\no&\no&\no&\no&\no\\
    \midrule
    \multicolumn{11}{c}{\textbf{Prompting} (zero-shot)} \\
    \midrule   
                                        & \#P       & en    & de    & es    & zh    & \textbf{avg}       & en & de & es & zh & \textbf{avg} \\
    \midrule
    gte-Qwen2-7B-instruct               &  7.61B    & 0.624   & 0.408   & 0.473   & 0.565   & \textbf{0.518}  & 0.537 & 0.496 & 0.442 & 0.495 & \textbf{0.493} \\
    Linq-Embed-Mistral                  &  7.11B    & 0.583   & 0.352   & 0.431   & 0.512   &  0.470          &0.489  &  0.467 & 0.466 & 0.487 & 0.477\\
    m-e5-large-instruct                 &  560M     & 0.532   & 0.368   & 0.422   & 0.535   & 0.464           &\no&\no&\no&\no&\no\\
    \bottomrule
    \end{tabular}
\end{table}

\footnotetext[2]{We only obtained these results after the deadline, hence we submitted the m-e5-large-instruct model for the test data.}

\begin{table}
    \centering
    \caption{\textbf{Cross-lingual Task A Test Results.}  We report mean-average precision (MAP) for each run on all fully in-language pairs (en–en, es–es, de–de, zh–zh) and on the three cross-language pairs that involve English (en–es, en–de, en–zh).  The macro average across English, Spanish, and German is given in the second column.}
    \label{tab:crosslingual}
    \setlength{\tabcolsep}{0.45em}
    \begin{tabular}{l c c c c c c c c}
        \toprule
        \multirow{2}{*}{} & \multirow{2}{*}{Avg.\,(en,es,de)} & \multicolumn{4}{c}{\textbf{In-language}} & \multicolumn{3}{c}{\textbf{Cross-lingual}}\\
        \cmidrule(lr){3-6}\cmidrule(l){7-9}
        & & en–en & es–es & de–de & zh–zh & en–es & en–de & en–zh\\
        \midrule
        TalentCLEF Baseline  & 0.340 & 0.408 & 0.324 & 0.348 & 0.380 & 0.335 & 0.345 & \no\\
        \midrule
        gte-Qwen2-7B-instruct  & \textbf{0.493} & \textbf{0.537} & \textbf{0.496} & 0.442 & \textbf{0.495} & \textbf{0.492} & 0.461 & \textbf{0.494}\\
        Linq-Embed-Mistral & 0.477 & 0.489 & 0.467 & 0.466 & 0.487 & 0.455 & 0.468 & 0.454\\
        m-e5-large (contrastive)  & 0.480 & 0.511 & 0.446 & \textbf{0.484} & 0.477 & 0.434 & \textbf{0.469} & 0.480\\
        \bottomrule
    \end{tabular}
\end{table}

\section{Results}

\subsection{Task A: Multilingual Job Title Matching}

For task A, the contrastive models achieves the highest performance on the validation data, and prompting on the test data, but their results are quite close on bath datasplits. It should be noted that the prompting model is 14 times larger, but did not require fine-tuning. The classification based models perform worse. The trends across different language models are consistent, larger models perform better, and m-e5-large provides a strong performance across approaches (except for prompting, due to its size). Performance on the languages also shows a highly similar trend. English gets the best performance, followed by Chinese, Spanish, and German. This is somewhat surprising, as Chinese is more distant to the other languages, and also completely unseen during training. 
We also note that the performance on the test data is lower compared to the validation data for all models. This could be a sign of overfitting, but after inspecting the data, we also saw that there is a larger overlap of job titles (\textasciitilde20\% versus \textasciitilde0\%) with the train data.
We hope details about the data creation process can shed more light on these differences. 


 \begin{table}
    \centering
    \caption{\textbf{Results Task B.} We show three methods for prediction; classification, similarity, and prompting-based methods. For the test predictions, we select the best performing model from each category.}
    \label{tab:b}
    \footnotesize
   \begin{tabular}[t]{m{7cm} m{7cm}}
    \vspace{-.4cm}
    \begin{tabular}{ll ll}
    \toprule   
    && \multicolumn{1}{l}{\textbf{Valid.}} & \multicolumn{1}{l}{\textbf{Test}} \\
    \midrule
    \multicolumn{4}{c}{\textbf{Classification} (fine-tuned)} \\
                                        &  \#P      & en & en \\
    \midrule
TalentCLEF                    & ?    & 0.187 & 0.196\\
\midrule
escoxlm-r                     & 561M  & 0.255  & \no\\
mdeberta-v3-base              & 276M  & \textbf{0.267} & \textbf{0.290}\\
m-e5-large-instruct           & 560M  & 0.266 & \no\\
m-e5-large                    & 560M  & 0.264 & \no\\
par-m-mpnet-base-v2           & 278M  & 0.247 & \no\\
    \bottomrule
    \end{tabular} &     \vspace{.3cm}
    \begin{tabular}{ll ll}
    \toprule
        && \multicolumn{1}{l}{\textbf{Valid.}} & \multicolumn{1}{l}{\textbf{Test}} \\
    \midrule
    \multicolumn{4}{c}{\textbf{Contrastive} (fine-tuned)} \\
    \midrule
                                           & \#P & en & en \\
    \midrule
    escoxlm-r                    & 561M  & 0.204  &  \no\\
    m-e5-large                   & 560M  & 0.214  &  \no\\
    par-m-mpnet-base-v2          & 278M  & \textbf{0.250}  & 0.253\\
    \midrule
    \multicolumn{4}{c}{\textbf{Prompting} (zero-shot)} \\
    \midrule
                                    & \#P    & en & en \\
    \midrule
gte-Qwen2-7B-instruct               &  7.61B    & \textbf{0.222} & 0.283\\
    Linq-Embed-Mistral                  &  7.11B    & 0.209 & \no\\
    m-e5-large-instruct                 &  560M     & 0.178 & \no\\
    \bottomrule
    \end{tabular}
    \end{tabular}
     \end{table}

Table~\ref{tab:crosslingual} breaks the scores down by language pair for the prompt-based models (which are the only one we submitted for the crosslingual track). \texttt{gte-Qwen2-7B-instruct} achieves the strongest English–Spanish (0.492) and English–Chinese (0.494) transfer, and ties for the best English–German performance (0.461). In-language results show the same trend: It tops English (0.537) and Spanish (0.496), while remaining competitive on German (0.442).  These findings suggest that (i) prompting benefits most from the larger pre-training signal in English and Spanish ESCO descriptions, and (ii) cosine-similarity scoring is robust across both monolingual and cross-lingual retrieval scenarios.

\subsection{TaskB: Job Title-Based Skill Prediction}
For task B (Table~\ref{tab:b}) the performance of the models is swapped. The classification based models outperform the other models on both the validation and the test data. Overall, the MAP scores are also much lower compared to task A, showing that the mapping of skills to jobs is a more challenging task. Therefore, we hypothesize that the direct supervised training signal is the key to the higher performance. Interestingly, models size has a less clear impact compared to task A, for both the contrastive and the classification models a the smallest language model performs best. 

\section{Analysis}

\paragraph{Per-category Performance.} For analysis, we investigate in which ESCO job title major group the models perform best for Task A. We take the three best-performing models in each method category (i.e., classification, contrastive, prompting) from Task A and map each data point from the validation set to their respective ESCO major group. In Table~\ref{tab:coverage}, we report the amount of job titles which can be mapped to a specific ESCO code. For English, there are 77.4\% unmapped titles and for both Spanish and German this is around 87\%. 

We report the results on the mapped job titles in Table~\ref{tab:map-major-groups-all}. We observe that for all models, job titles in categories such as ``managers'', ``professionals'', ``technicians and associate professionals'', ``clerical support workers'', and ``service and sales workers'' are the most difficult to predict. In contrast, we see job titles from categories such as ``armed forces occupations'', ``craft and related trades'', ``plant and machine operators'' and ``elementary occupations'' being often predicted correctly. In the case of unmapped job titles, we see that most models do not perform well. 

\begin{table}[t]
  \centering
  \caption{Task A: Corpus mapping coverage by language in the validation set.}
  \label{tab:coverage}
  \small
  \begin{tabular}{lrrr}
    \toprule
    & \textbf{English} & \textbf{Spanish} & \textbf{German} \\
    \midrule
    Total corpus titles   & 2619 & 4661 & 4729 \\
    Mapped titles         &  591 &  569 &  585 \\
    Unmapped titles (\%)  & 77.4\% & 87.8\% & 87.6\% \\
    \bottomrule
  \end{tabular}
\end{table}


\begin{table}[t]
  \centering
  \caption{Mean average precision score by ESCO major group across languages and model variants.}
  \label{tab:map-major-groups-all}
  \scriptsize
  \begin{tabular}{l
                  *{3}{ccc}}
    \toprule
    & \multicolumn{3}{c}{\textbf{English}} 
    & \multicolumn{3}{c}{\textbf{Spanish}} 
    & \multicolumn{3}{c}{\textbf{German}} \\
    \cmidrule(lr){2-4}\cmidrule(lr){5-7}\cmidrule(lr){8-10}
    
      & \textit{clas.} & \textit{contr.} & \textit{prompt}
      & \textit{clas.} & \textit{contr.} & \textit{prompt}
      & \textit{clas.} & \textit{contr.} & \textit{prompt} \\
       \textbf{ESCO major group} 
      & \textit{E5-inst} & \textit{E5-con} & \textit{GTE} 
      & \textit{E5-inst} & \textit{E5-con} & \textit{GTE} 
      & \textit{E5-inst} & \textit{E5-con} & \textit{GTE} \\
    \midrule
    Unmapped Titles  
      & 0.553 & 0.551 & 0.617 
      & 0.433 & 0.478 & 0.477 
      & 0.431 & 0.462 & 0.426 \\
    0 Armed forces occupations  
      & 1.000 & 1.000 & 0.958 
      & 1.000 & 1.000 & 0.833 
      & 0.861 & 1.000 & 0.667 \\
    1 Managers                   
      & 0.788 & 0.840 & 0.823 
      & 0.553 & 0.595 & 0.633 
      & 0.582 & 0.669 & 0.591 \\
    2 Professionals              
      & 0.647 & 0.668 & 0.702 
      & 0.569 & 0.585 & 0.639 
      & 0.545 & 0.573 & 0.575 \\
    3 Technicians \& assoc.\ profs. 
      & 0.703 & 0.693 & 0.775 
      & 0.574 & 0.621 & 0.675 
      & 0.611 & 0.659 & 0.557 \\
    4 Clerical support workers  
      & 0.774 & 0.857 & 0.729 
      & 0.537 & 0.581 & 0.533 
      & 0.468 & 0.542 & 0.374 \\
    5 Service \& sales workers   
      & 0.764 & 0.728 & 0.729 
      & 0.772 & 0.961 & 0.898 
      & 0.698 & 0.707 & 0.531 \\
    6 Skilled agri./forestry/fishery workers
      & — & — & — 
      & — & — & — 
      & — & — & — \\
    7 Craft \& related trades     
      & 0.917 & 1.000 & 0.889 
      & 0.678 & 0.747 & 0.700 
      & 0.765 & 0.697 & 0.869 \\
    8 Plant \& machine operators  
      & 1.000 & 0.833 & 1.000 
      & 0.875 & 0.479 & 0.875 
      & 1.000 & 1.000 & 1.000 \\
    9 Elementary occupations      
      & 0.822 & 1.000 & 0.989 
      & 0.441 & 0.793 & 0.552 
      & 0.770 & 0.815 & 0.433 \\
    \bottomrule
  \end{tabular}
\end{table}

\section{Conclusion}
In this paper, we report our methods for the 2025 TalentCLEF shared task. We demonstrate that prompting is effective for multilingual job title matching (Task A) and classification approaches excel in predicting job-related skills (Task B). However for task B, prompting-based methods, despite their lower performance, show promising results in a zero-shot scenario, suggesting potential avenues for further exploration.

\begin{acknowledgments}
MZ is supported by the research grant (VIL57392) from VILLUM FONDEN.
\end{acknowledgments}

\section*{Declaration on Generative AI}
\noindent{\em During the preparation of this work, the author(s) used GPT-4o in order to: Check grammar and spelling check. After using these tool(s)/service(s), the author(s) reviewed and edited the content as needed and take(s) full responsibility for the publication's content. 

\bibliography{sample-ceur, anthology}

\appendix

\end{document}